\documentclass[letterpaper, 10 pt, conference]{ieeeconf}  

\IEEEoverridecommandlockouts                              

\usepackage{graphicx} 
\usepackage{amsmath} 
\usepackage{amssymb}  
\usepackage{subfigure}
\usepackage{balance}

\makeatletter
\let\NAT@parse\undefined
\makeatother
\usepackage[numbers]{natbib}

\usepackage{irap_misc}
\usepackage{irap_acronyms}
\usepackage{irap_SIunits}

\usepackage{soul,color}
\usepackage{xcolor}

\title{\LARGE \bf
Deep Learning from Shallow Dives:\\Sonar Image Generation and Training for Underwater Object Detection
}

\author{Sejin Lee$^{1}$ and Byungjae Park$^{2}$ and Ayoung Kim$^{3}$
\thanks{$^{1}$Sejin Lee is with the Division of Mechanical \& Automotive Engineering,
        Kongju National University, 1223-24 Cheonan-daero, Cheonan 31080, Republic of Korea
        {\tt\small sejiny3@kongju.ac.kr}}%
\thanks{$^{2}$Byungjae Park is with the Intelligent Robot System Research Group,
	    ETRI, 218 Gajeong-ro, Yuseong-gu, Daejeon, 34129, Republic of Korea.
        {\tt\small bjp@etri.re.kr}}%
\thanks{$^{3}$Ayoung Kim is with the Department of Civil and Environmental Engineering, KAIST,
291 Daehak-ro, Yuseong-gu, Daejeon 34141, Republic of Korea
        {\tt\small ayoungk@kaist.ac.kr}}%
}

\begin{document}

\maketitle
\thispagestyle{empty}
\pagestyle{empty}

\begin{abstract}

Among underwater perceptual sensors, imaging sonar has been highlighted for its
perceptual robustness underwater. The major challenge of imaging sonar, however,
arises from the difficulty in defining visual features despite limited
resolution and high noise levels. Recent developments in deep learning provide a
powerful solution for computer-vision researches using optical images.
Unfortunately, deep learning-based approaches are not well established for
imaging sonars, mainly due to the scant data in the training phase. Unlike the
abundant publically available terrestrial images, obtaining underwater images is
often costly, and securing enough underwater images for training is not
straightforward. To tackle this issue, this paper presents a solution to this
field's lack of data by introducing a novel end-to-end image-synthesizing method
in the training image preparation phase. The proposed method present image
synthesizing scheme to the images captured by an underwater simulator. Our
synthetic images are based on the sonar imaging models and noisy characteristics
to represent the real data obtained from the sea. We validate the proposed
scheme by training using a simulator and by testing the simulated images with
real underwater sonar images obtained from a water tank and the sea.

\end{abstract}

\section{Introduction}
\label{sec:intro}


In many underwater operations \cite{cho-2015, purcell-2011, reed-2003,
belcher-2000, lee-2013, williams-2011, galceran-2012}, perceptual object
detection and classification are required, such as search and rescue, evidence
search, and defense missions for military purposes. Bodies of water often
present a critical decrease in visibility due to the high density of fine floats
or aquatic microorganisms \cite{lee-2017}. Due to this limitation of using
optical images, imaging sonar has been a widely accepted solution providing
reliable measurements regardless of the water's turbidity
\cite{yshin-2015-oceans, johnnsson-2010}. Although sonars extend the perceptual
range, the resulting images follow a different projection model, resulting in
less intuitive and low-resolution images and cannot be easily understood by
human operators. In addition, due to the sensor's physical characteristics, a
considerable level of noise is generated in the water image, so it is difficult
to ensure the reliability of sonar image analysis and identification
\cite{navigator}.

\begin{figure}[!t]
    \centering
    \includegraphics[width=0.95\columnwidth]{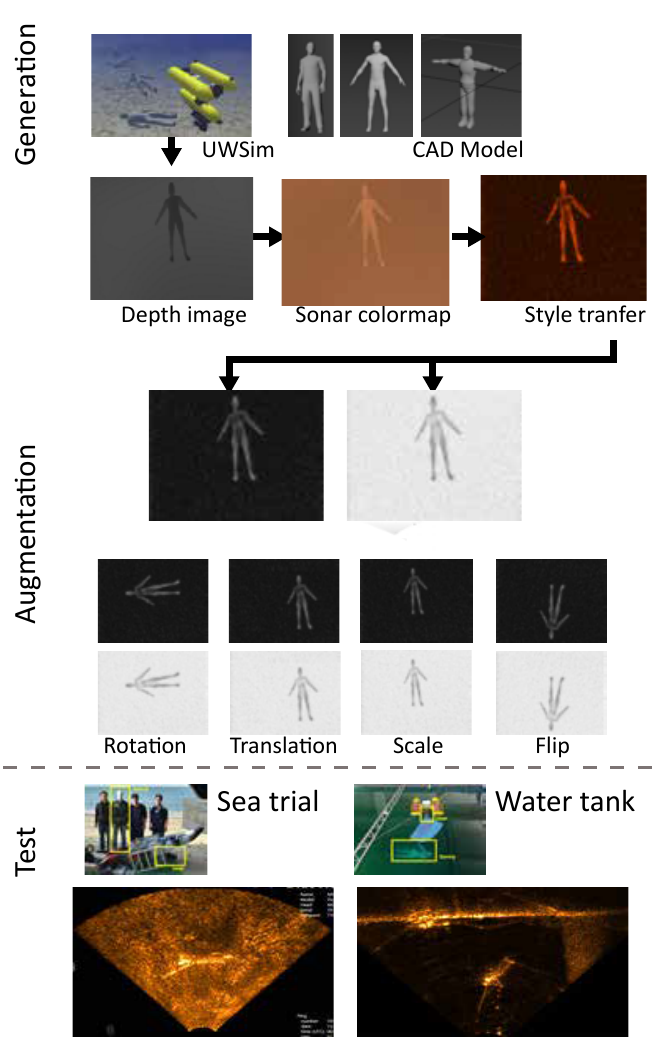}

    \caption{Overview of the proposed method. We propose a sonar image
    synthesizing for the images generated by a simulator. We trained using
    images captured and synthesized from simulator, and tested over real
    underwater target detection scenario in water tank and real sea.}

    \label{fig:overview}
\end{figure}

Early work on sonar image-based classification was aimed at  \ac{ATR} or
sediment classification. Low resolution and image ambiguity due to the shadowing
effect has always been an issue in defining handcrafted features for sonar
images. \citeauthor{enric-2012} used multibeam, forward-looking sonar to detect
man-made objects \cite{enric-2012}. In their work, they applied a series of
estimation modules to detect objects. In \cite{zhou-2004}, the authors employed
power spectral analysis methods for seafloor sediment classification.  In
\cite{williams-2015-tgrs}, the author reported a useful measure for sonar
imaging called \textit{lacunarity} to classify seafloor characteristics.


Recently, to overcome these challenges, deep-learning-based approaches have been
introduced. Researchers focused on a partial application. They exploited feature
learning by learning features from the \ac{CNN} and then piped the learned
feature into another machine learning algorithm like a \ac{SVM}
\cite{pzhu-2017-cdc}. Other researchers used deep learning in a more end-to-end
fashion.  In \cite{williams-2016-icpr}, target classification using \ac{SAS}
images was introduced. Two target objects were considered in the study, and the
performance of the CNN-based approach was compared to that of the typical
feature-based classifier. \citeauthor{jkim-syu-2016-oceans} also focused on
applying deep learning for vehicle detection using forward-looking
sonar~\cite{jkim-syu-2016-oceans}.  In the aforementioned approaches, however,
the authors collected real sonar images and divided them into training and test
image sets.

When applying deep learning based approaches in underwater environment, training
with real sonar images from the target environment would be optimal but is
highly challenging in several aspects. First, the underwater imaging
specifically for classification results in a biased dataset. Second, obtaining
underwater images demands time and effort. Applying deep learning underwater
addresses the major challenge of scanty data. Many efforts have been made to
alleviate the training data shortage. For example, \cite{jmckay-2017-oceans}
exploited existing pre-trained weights from in-air images. They applied fine
tuning using sonar images.

A similar strategy to ours recently found in the literature synthetically
generates photo-realistic images. Authors in \cite{kdenos-2017-oceans} examined
the synthetic training set generation by applying a proper background of white
noise to the simulated images. This synthetic training image generation was also
thoroughly handled in \cite{jchen-2017-jasa}. The authors evaluated the \ac{GAN}
to learn underlying features in an unsupervised manner. They also examined the
effect of style transfer on background and shadow generation ability.


Differing from those early studies who focused on generation of images, this
paper proposes an end-to-end solution to prepare a training dataset for
underwater object detection and validating with real underwater sonar images. In
\figref{fig:overview}, we present a simulator-based training data generation
tool specifically for underwater sonar images. Our contributions are as follows:

\begin{itemize}

\item We propose a solution to the problem of scanty data in underwater sonar
applications by proposing synthetic training image generation via style
transfer. The proposed method takes one channel depth image from a simulator to
provide various aspects (e.g., scale, orientation and translation) of the
captured data.

\item We performed a thorough evaluation using real sonar data from pool and sea
trial experiments to validate the proposed method. Specifically, we present that
the proposed simulation trained network performs equally well as the real sea
data trained network. By doing so, the proposed training scheme alleviates the
training data issue in underwater sonar imaging studies.

\item We also verified the trained network with sample images from various sonar
sensors. The test sonar images are sampled from video provided by sonar
companies. This validation proves the proposed scheme could be widely applicable
for sonar images captured from various underwater environment. Note that the
sonar data used in testing was never been used in training phase. Therefore, we
suggest elimination of the real-data acquisition phase in deep learning for
underwater application.

\end{itemize}

\section{Training Set Generation}
\label{sec:sim}

In this section, we introduce simulation-created training data generation for
underwater object detection.

\subsection{Base Images Preparation from Simulator}

Obtaining real images from the ocean would be ideal, as reported in
\cite{williams-2016-icpr}, where the author collected eight years of data from
marine missions to prepare and test classifications.  However, as reported, data
collection in underwater missions is demanding. To overcome this limitation, we
captured a base image for the synthetic training dataset from a simulated depth
camera in the UWSim \cite{uwsim}. Using a simulator allowed us to train with
various objects by loading a 3D CAD model of the target objects. By diversifying
pose and capturing altitude, multiple scenes of objects were collected, as shown
in the sample scene in \figref{fig:overview}.


The UWSim provides a diverse choice of underwater sensor modalities, and users
may be able to implement their own sensor module within the simulator
\cite{dgwon-2017-urai}. Developing a detailed sensor module for a specific
perceptual sensor would require careful design of the modules in the simulator
based on the detailed understanding of the sensor and the environment. However,
we found that generating a photo-realistic image from a rather simple depth
image may provide a feasible solution. Specifically, we proposed using the style
transfer to generate realistic-enough synthetic sonar images for training, and
the simulator only needed to provide a basic representation of the scene for
style transfer to be applied.

As depicted in \figref{fig:overview}, using the captured depth image from the
simulator, we applied a colormap in \cite{dgwon-2017-urai} added by white noise.
Then the images were normalized and prepared as a base image before entering the
style transfer phase.


\subsection{Image Synthesizing}

\begin{figure}[!h]
  \centering
  \subfigure[StyleBankNet architecture]{%
    \includegraphics[width = 0.95\columnwidth]{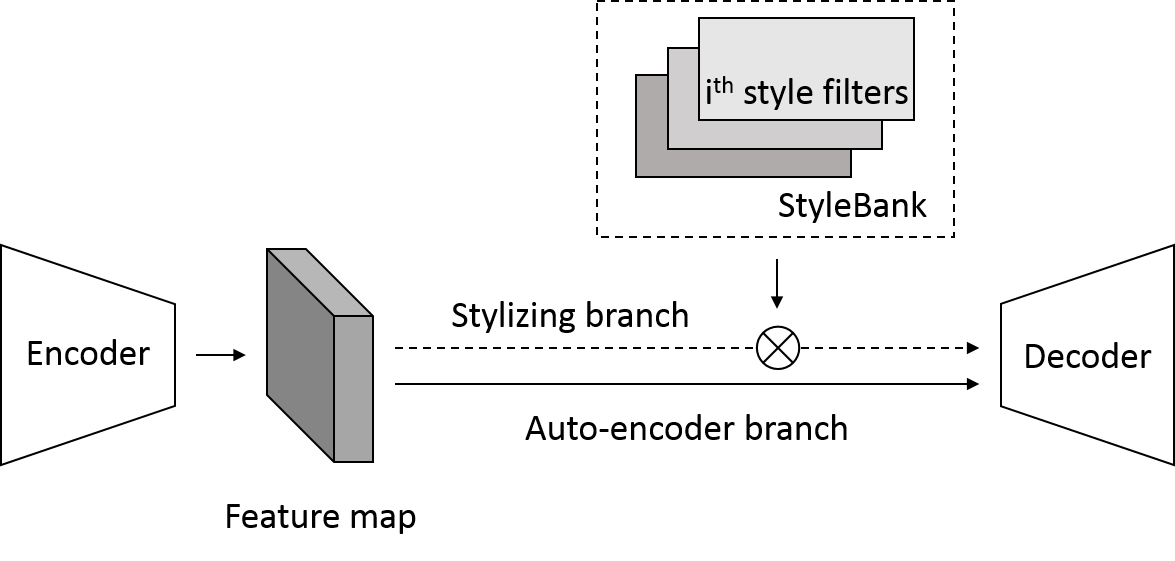}
    \label{fig:stylebanknet_architecture_overview}
  }
	\subfigure[Detailed architecture of encoder, decoder, and style filters in StyleBank]{
	  \label{tab:stylebank_architecture_detailed}
			\centering
	    \scriptsize
      \vspace{10px}
      \begin{tabular}{c | c}
      Name & Architecture  \\
      \hline\hline
      Encoder & $c9s2-32, IN, C64, IN, C128, IN, C256, IN$ \\
      Decoder & $TC128, IN, TC64, IN, C32, IN, tc9s2-3$ \\
      $i^{\text{th}}$ style filters in SB & $C256, IN, C256, IN$
      \vspace{5px}
      \end{tabular}
	    \normalsize
	}

\caption{Network architecture of StyleBankNet. It consists of three modules: encoder, decoder and StyleBank (SB).  $c9s2-32$: 9$\times$9 convolutional block with 32 filters and stride 2, $IN$: Instance Normalization, $Cn$: 3$\times$3 convolutional blocks with $n$ filters and stride 1, $TCn$: 3$\times$3 transposed convolutional blocks with $n$ filters with stride 1, and $tc9s2-3$: 9$\times$9 transposed convolutional block with 3 filters and stride 2.}
\label{fig:stylebank_architecture}
\end{figure}

Given this base image, we adopted the StyleBankNet \cite{stylebank} to
synthesize the noise characteristics of sonar images acquired in various
underwater environments, such as water tank and sea. This network simultaneously
learns multiple target styles using an encoder ($\mathcal{E}$), decoder
($\mathcal{D}$), and StyleBank ($\mathcal{K}$), which consists of multiple sets
of style filters (\figref{fig:stylebank_architecture}). Each set of style
filters represents the style of one underwater environment.

\subsubsection{Losses}

There are two different branches in the StyleBankNet: auto-encoder branch
($\mathcal{E} \rightarrow \mathcal{D}$) and stylizing branch ($\mathcal{E}
\rightarrow \mathcal{K} \rightarrow \mathcal{D}$). The StyleBankNet uses these
branches to decouple styles and contents of sonar images. The auto-encoder
branch uses a reconstruction loss to train the encoder and decoder for
generating an output image that is as close as possible to an input image.
\begin{align}
\mathcal{L}_{\mathcal{R} } (\it{C}, \it{O}  ) = || \it{O} - \it{C} ||^{2},
\label{eqn:reconstruction_loss}
\end{align}
where $C$ and $O$ is input and output images, respectively.  The stylizing
branch uses a {\it{perceptual loss}}  to jointly train the the encoder, decoder,
and StyleBank \cite{jjohnson-2016-eccv}:
\begin{align}
\nonumber
\mathcal{L}_{\mathcal{P} } (\it{C}, \it{S}_{i}, \it{O}_{i} ) &= \alpha \cdot \mathcal{L}_{c} (\it{O}_{i}, \it{C_i} )
\nonumber
\\ & ~ + \beta \cdot \mathcal{L}_{s} (\it{O}_{i} , \it{S}_{i})
\nonumber
\\ & ~ + \gamma \cdot \mathcal{L}_{reg} (\it{O}_{i} )
\nonumber
\\ & ~ + \delta \cdot \mathcal{L}_{atki} (\it{O}_{i} , \it{S}_{i}),
\label{eqn:perceptual_loss}
\end{align}
where $\it{S}_{i}$ is one of images with $i^{\text{th}}$ style. $\mathcal{L}_{c}
(\it{O}_{i}, \it{C}_{i} )$, $\mathcal{L}_{s} (\it{O}_{i} , \it{S}_{i})$,
$\mathcal{L}_{reg} (\it{O}_{i} )$ and $\mathcal{L}_{atki} (\it{O}_{i} ,
\it{S}_{i})$ are feature reconstruction loss, style reconstruction loss,
regularization loss, and \ac{ATKI} loss, respectively \cite{jjohnson-2016-eccv}.
In this equation, the style reconstruction loss measures the difference between
output and style images in style such as colors, textures, patterns, etc:
\begin{align}
\mathcal{L}_{s} (\it{O}_{i}, \it{S}_{i} )= \sum_{l \in \{l_{s} \} } || G(F^{l}(\it{O}_{i}) ) - G(F^{l}(\it{S}_{i} ) ) ||^{2},
\label{eqn:style_loss}
\end{align}
where $F^{l}$ and G are feature map and Gram matrix computed from $l^{\text{th}}$ layer
of VGG-16 layers $l_{s}$, respectively.

The last term $\mathcal{L}_{atki} (\it{O}_{i} , \it{S}_{i})$ is a \ac{ATKI}
loss. We added it to the original perceptual loss \cite{jjohnson-2016-eccv}, as
the StyleBankNet is able to learn the unique intensity distribution
characteristics of sonar images. In a sonar image, some parts appear much
brighter than other parts. These brighter parts may contain objects of interest
because sonar signals are reflected by objects and floors.  Although the
intensity distribution of the brighter parts is much different from the global
intensity distribution, it is likely to be overlooked when computing the style
reconstruction loss to train the StyleBankNet because the brighter parts are
usually much smaller than other parts. As a result, the characteristics of the
brighter parts are not learned appropriately. Motivated by the \ac{ATKI}
\cite{yfan-2017-nips}, the \ac{ATKI} loss is used to measure the intensity
distributions of brighter parts in output and style images.
\begin{align}
\mathcal{L}_{atki} (\it{O}_{i}, \it{S}_{i} )= \frac{1}{k} \sum^{k}_{j=1} || \it{O_{G} }_{i}^{[j]}  - \it{S_{G} }_{i}^{[j]} ||^{2},
\label{eqn:atki_loss}
\end{align}
where $\it{O_{G} }_{i}^{[j]}$ and $\it{S_{G} }_{i}^{[j]}$ are the $j^{\text{th}}$ largest intensity values in grayscale output and style images, respectively. By applying the \ac{ATKI} loss, the unique intensity distributions can be synthesized by the StyleBankNet.

%
%
%

\begin{figure}[!t]
	\centering
	\includegraphics[width = 0.95\columnwidth]{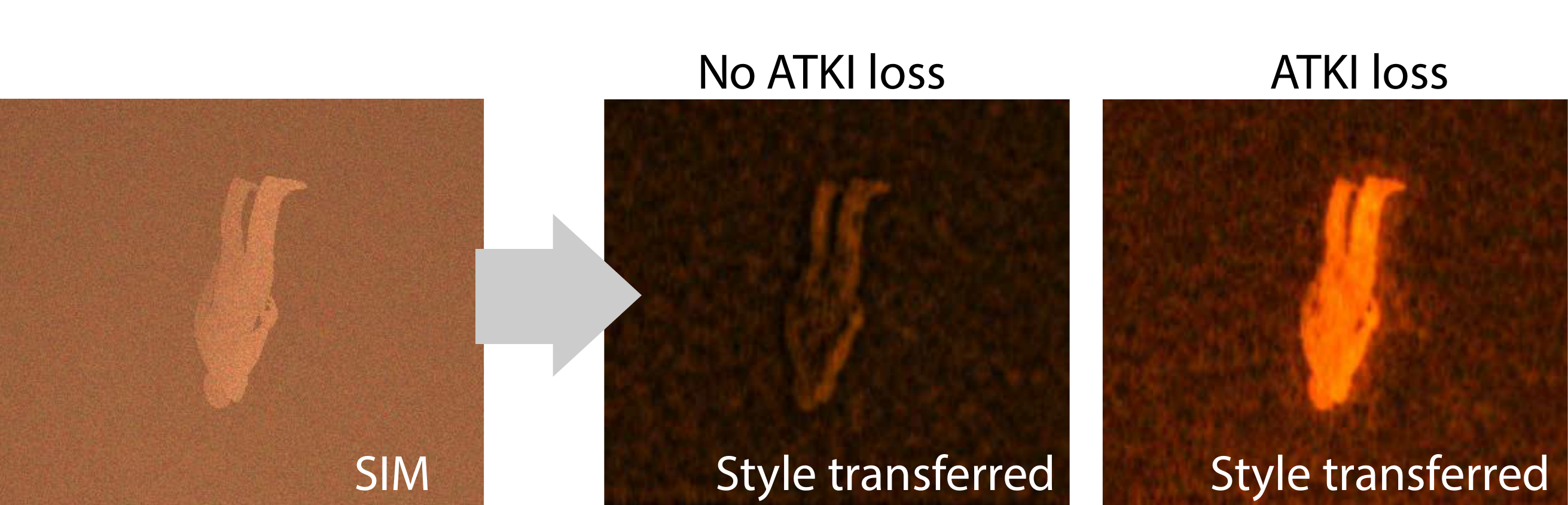}

  \caption{Effect of \ac{ATKI}. By considering additional loss from \ac{ATKI}, the target object is styled more strongly.}

	\label{fig:atkiloss}
\end{figure}

The effect of \ac{ATKI} loss is depicted in \figref{fig:atkiloss}.  When
additionally using the \ac{ATKI} loss, the target object appears more clearly
and brightly than when not using the loss.

\subsubsection{Training}

The dataset for training the StyleBankNet consists of a content set, which is
composed of base images and multiple style sets (e.g., pool and sea).  Each set
contains object-centered 300 images. A single mini-batch consists of randomly
sampled content images and style images with style indices. To better examine
generalized characteristics of sonar images, a pair between base and style
images is not fixed in each iteration.  A (${T+1}$)-step alternative training
strategy is employed to ensure a balanced learning of the encoder, decoder, and
StyleBank using two branches \cite{igoodfellow-2014-nips}. Parameters of the
StyleBankNet is updated in every ${T+1}^{\text{th}}$ iteration using the
auto-encoder branch; otherwise, the styling branch is used.

\section{Application to Object Detection}

\subsection{\ac{CNN} Architecture}

We used the deep learning toolbox including the Faster \ac{R-CNN}
\cite{ren2017faster} model released in Matlab for underwater object detection.
Although the region proposal algorithms such as EdgeBoxes \cite{zitnick2014edge}
or Selective Search \cite{uijlings2013selective} are typically applied, the use
of these techniques becomes the processing bottleneck in the older model
\cite{girshick2015fast}.  Faster \ac{R-CNN} addresses this issue by implementing
the region proposal mechanism using the CNN and thereby making region proposal
in the CNN training and prediction steps. For this \ac{RPN} training, the layers
were basically set up as follows; Input layer ($32\times32\times3$), 1st
Convolution layer ($5\times5\times32$), Relu, MaxPooling ($3\times3$), 2nd
Convolution layer ($3\times3\times64$), Relu, MaxPooling($3\times3$), 3rd
Convolution layer($3\times3\times32$), Relu, MaxPooling($3\times3$),
Fully-connected Layer(200), Relu, Fully-connected Layer(2), Softmax Layer,
Classification layer. In some training cases, the hyper-parameters were slightly
reduced compared to this base model.

\subsection{Training Image Augmentation}

The style transferred image can be directly used for training, and the
synthesized images themselves could be sourced for many sonar imaging
applications. In this application, we propose a synthesizing scheme generally
applicable to various sonar images. Thus, in this augmentation phase, we
converted the images to grayscale and their inverted image in the form of
general one-channel sonar images. Including inverted images is critical for
sonar images because when an object is imaged by sonar, the intensity of the
object may be brighter or darker than the background depending on the relative
material property of the object and environment. To remedy this situation, we
generated two types of images from a single channel synthesized image, as shown
in \figref{fig:overview}.

For deep learning application, ensuring sufficient diversity in training
datasets is meaningful. When capturing data from the simulator, physical
diversity was considered to include various rotation, translation, and scaling.
Additionally, we randomly flipped the captured base images. We applied
variations in scale, rotation, and translation for the training dataset.

\section{Experimental Results}
\label{sec:results}

In this section, we provides a series of experiments to evaluate style transfer
performance and its application to object detection.


\begin{figure}[!t]
  \centering
  \subfigure[Water tank test (\texttt{POOL})]{%
    \includegraphics[width = 0.45\columnwidth]{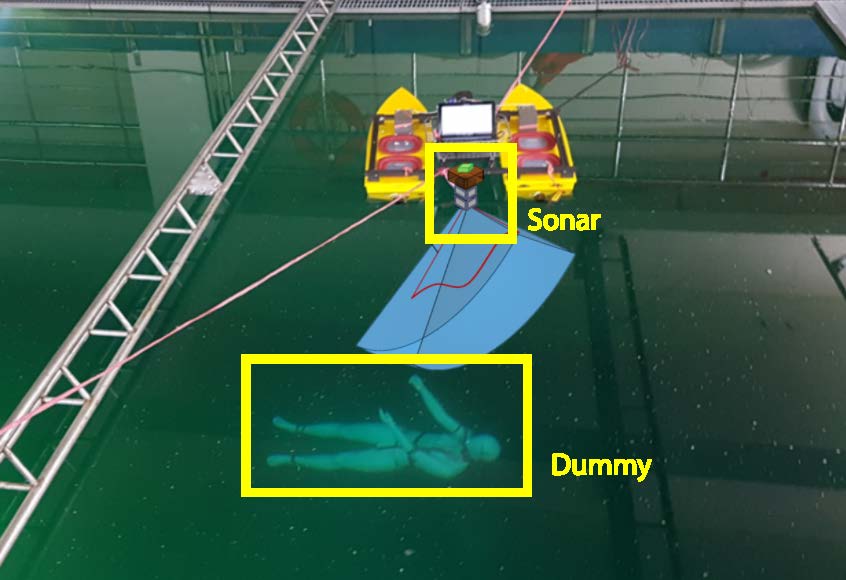}
    \label{POOL_picture}
  }
  \subfigure[Sea trial (\texttt{SEA})]{%
    \includegraphics[width = 0.43\columnwidth]{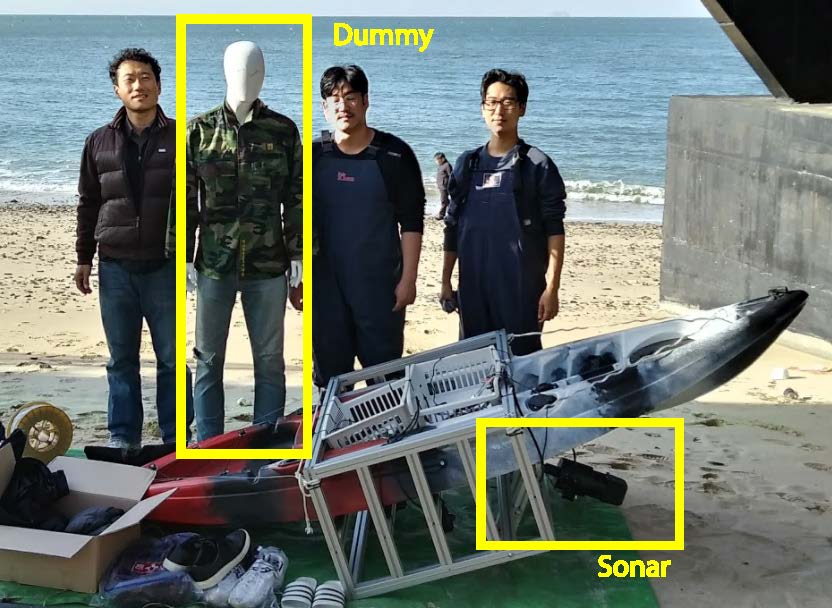}
    \label{SEA_picture}
  }
	\subfigure[Our own validation datasets]{
	  \label{tab:datalist}
	  \begin{minipage}[t]{0.9\columnwidth}
			\centering
		\vspace{10px}
	  \scriptsize
	  \begin{tabular}{c|ccc}
		  Name                 & Environment  & Description              & \# of Images\\
		  \hline\hline
		  \texttt{SIM}	       & UWSim        & Simulated depth camera   & 370\\
      \texttt{SIM-POOL}    & UWSim        & Water tank styled images & 370\\
      \texttt{SIM-SEA2017} & UWSim        & Sea styled images        & 370\\
      \hline
		  \texttt{POOL}	   	& Water tank   & Multibeam sonar images   & 735\\
		  \texttt{SEA2017}	& Sea          & Multibeam sonar images   & 1045\\
		  \hline
		  \texttt{SEA2018}	& Sea          & Multibeam sonar images   & 1935
      \vspace{5px}
	\end{tabular}
	\normalsize
\end{minipage}
	}

  \caption{Experiment set up for clean water tank (\texttt{POOL}) and real sea
  data (\texttt{SEA}). Sonar was mounted either on \ac{USV} (for \texttt{POOL})
  or kayak (for \texttt{SEA}).}

  \label{fig:exp_setup}
\end{figure}

\subsection{Datasets}

For training, images (\texttt{SIM}) are prepared using the algorithm described
in \secref{sec:sim}. The \texttt{SIM} images are styled targetting either water
tank (\texttt{SIM-POOL}) and sea (\texttt{SIM-SEA2017}) respectively. Details are
summarized in the table in \figref{fig:exp_setup}.

For validation, we used our own dataset listed in the table in
\figref{tab:datalist} together with the publicly available sample images from
sonar companies. When collecting our own validation dataset, images were
captured by imaging a human-sized dummy using a Teledyne BlueView M900-90, a
multibeam imaging sonar with a 90$^{\circ}$ field of view, 20$^{\circ}$ beam
width and \unit{100}{m} maximum range. Data were collected from a water tank and
from the sea as shown in \figref{fig:exp_setup}.

\begin{figure}[!t]
	\centering
	\includegraphics[width = 0.9\columnwidth]{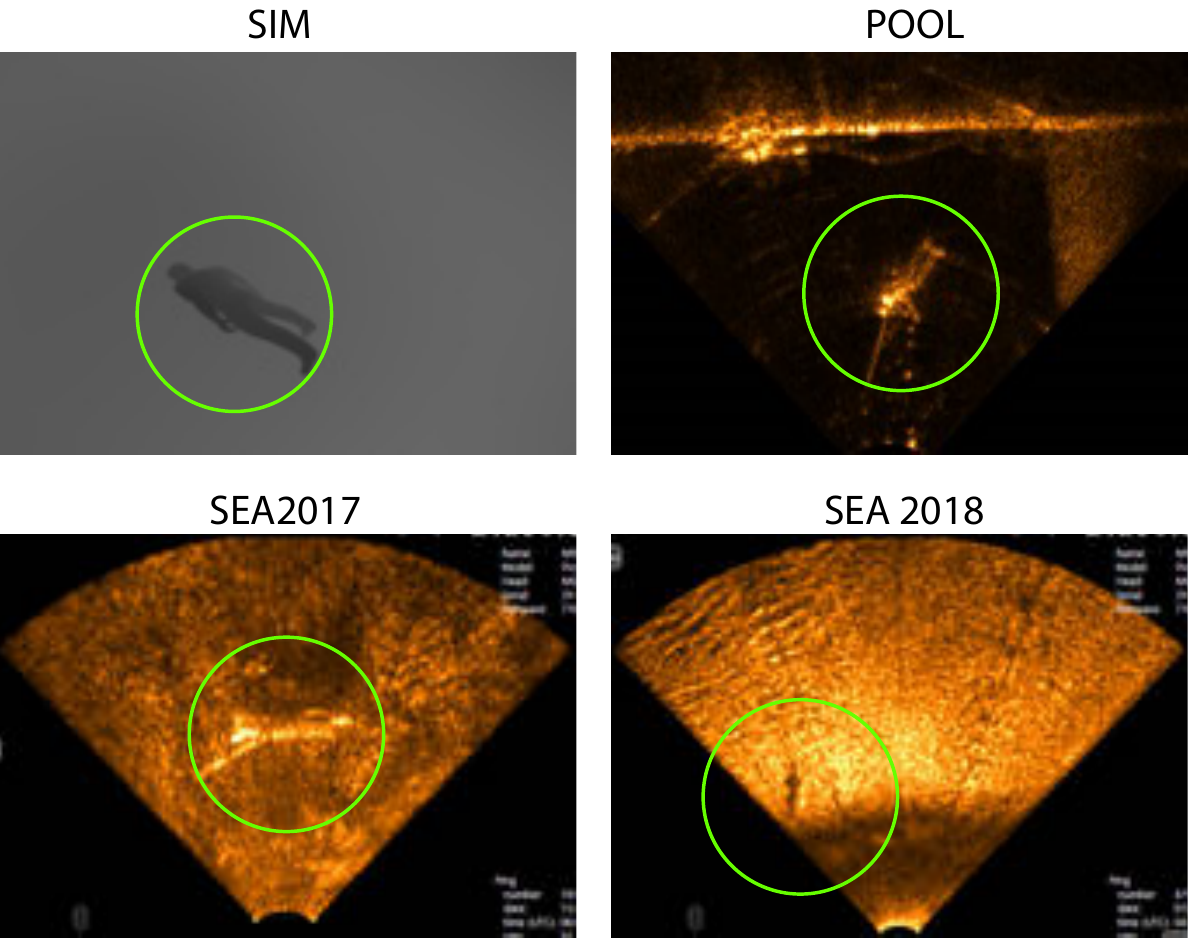}

  \caption{Sample raw images from each environment without applying style
  transfer. All four sample images contain the target object marked as a green
  circle.}

	\label{fig:samples}
\end{figure}

\begin{figure}[!b]
	\centering
	\includegraphics[width = 0.95\columnwidth]{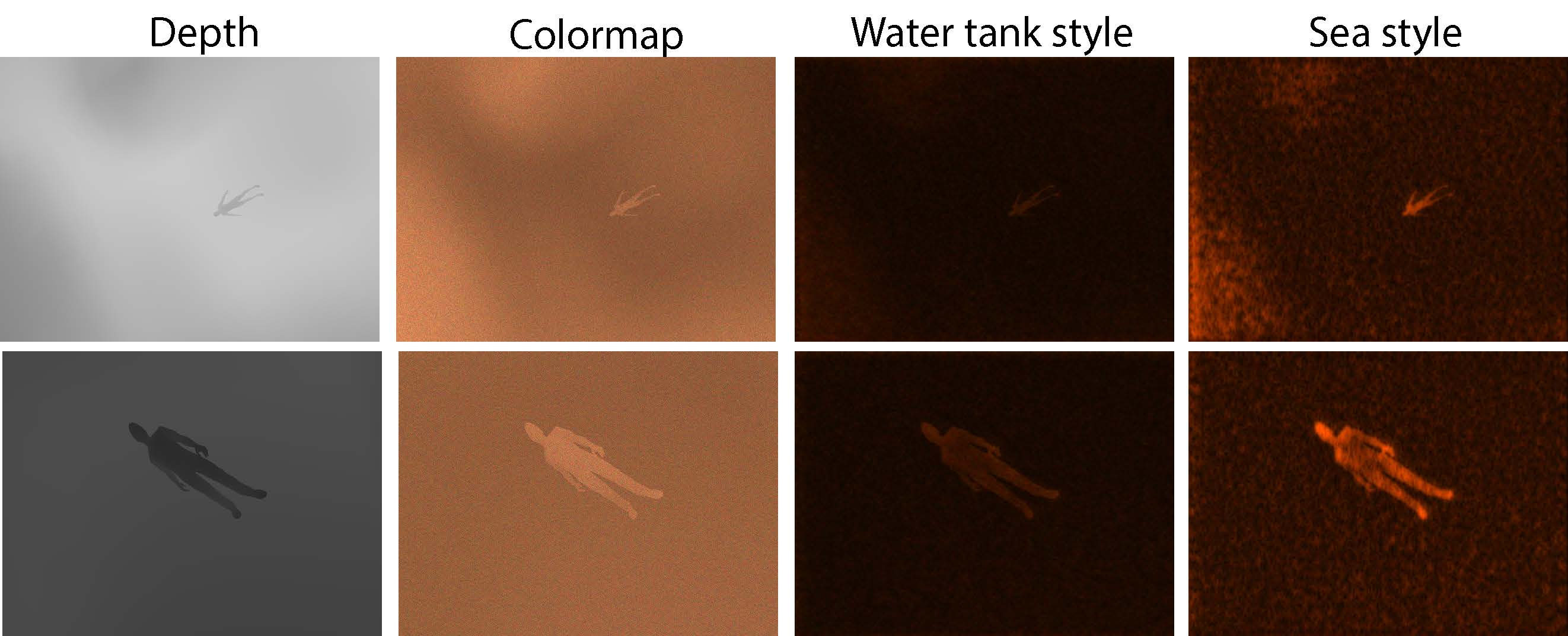}

  \caption{Style transferred image samples. Given the depth images captured from
  the simulator, we generate color-map changed images. On the third and fourth
  column, style transferred images are shown. When style transferred to the
  water tank, the images showed a darker background well representing the actual
  images captured in the water tank.}

  \label{fig:style}
\end{figure}

The first dataset, called \texttt{POOL}, was captured in the very clean water
testbed of the \ac{KIRO}. The maximum depth of this water testbed was
approximately \unit{10}{m}. The dummy was positioned in a water depth of about
\unit{4}{m} to simulate a submerged body, as shown in \figref{POOL_picture}. The
imaging sonar was mounted on a \ac{USV} platform that was capable of rotating
the sonar sensor at an angular interval of 5$^{\circ}$ and enabled collection of
underwater sonar images from various angles. The second dataset (\texttt{SEA})
was captured in severely turbid water from Daecheon Beach in Korea. The BlueView
M900-90 was fixed to the lower part of the kayak, as shown in
\figref{SEA_picture}, and the heading direction was mounted at about
30$^{\circ}$ downward from the water's surface. In this experiment, the distance
between the sensor and the dummy was about 2 to 4 meters. The \texttt{SEA}
dataset was collected twice, and the datasets were named \texttt{SEA2017} and
\texttt{SEA2018}. The dataset of \texttt{POOL}, \texttt{SEA2017} and
\texttt{SEA2018} has 735, 1045, and 1935 images containing a submerged body
respectively.

\figref{fig:samples} illustrates the sample images captured from each
environment. Images from water tank reveals relatively darker images than sea
trial. The appearance of the target object change drastically even when captured
in the same environment depending on the viewpoint and nearby sediment
condition. As can be seen \texttt{SEA2018} presents a brighter image than
\texttt{SEA2017} capturing the target object farther than previous data.

\subsection{Experimental Setup and Evaluation criteria}

\begin{figure}[!t]
	\centering
	\includegraphics[width = 0.95\columnwidth]{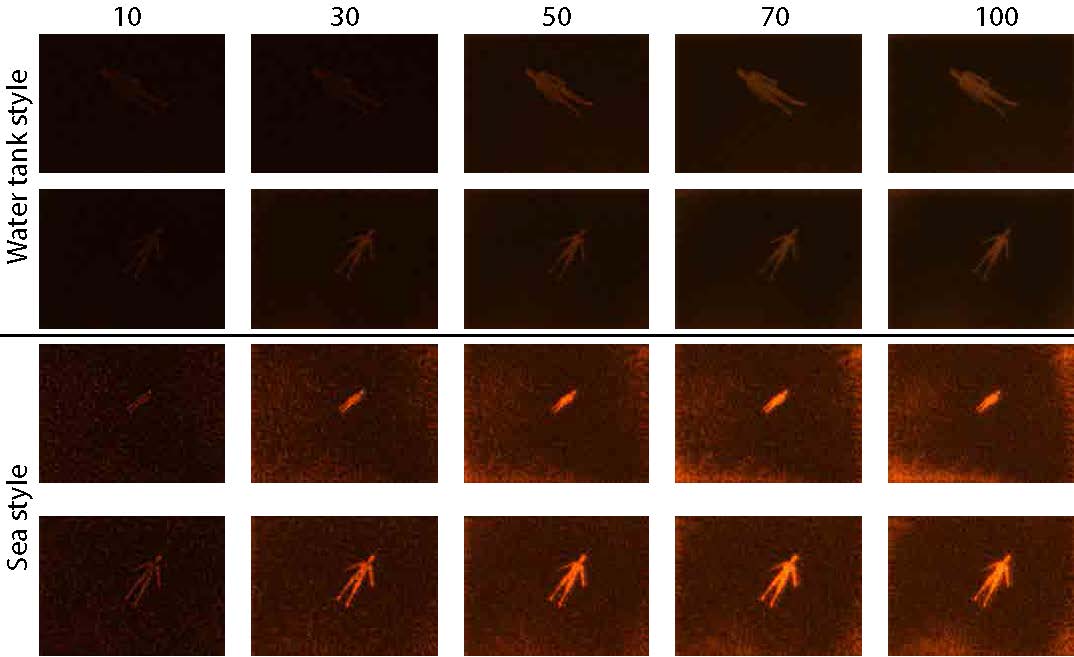}
  \caption{As the epoch evolves, the target object appears more clearly. At
  around 100 epoch, the shape of the body clearly shows and background reveals
  similar characteristics to the real images.}
  \label{fig:style_epoch}
\end{figure}

\begin{figure}[!b]
	\centering
  \subfigure[\texttt{SIM-POOL} to \texttt{POOL}]{
    \includegraphics[width = 0.45\columnwidth]{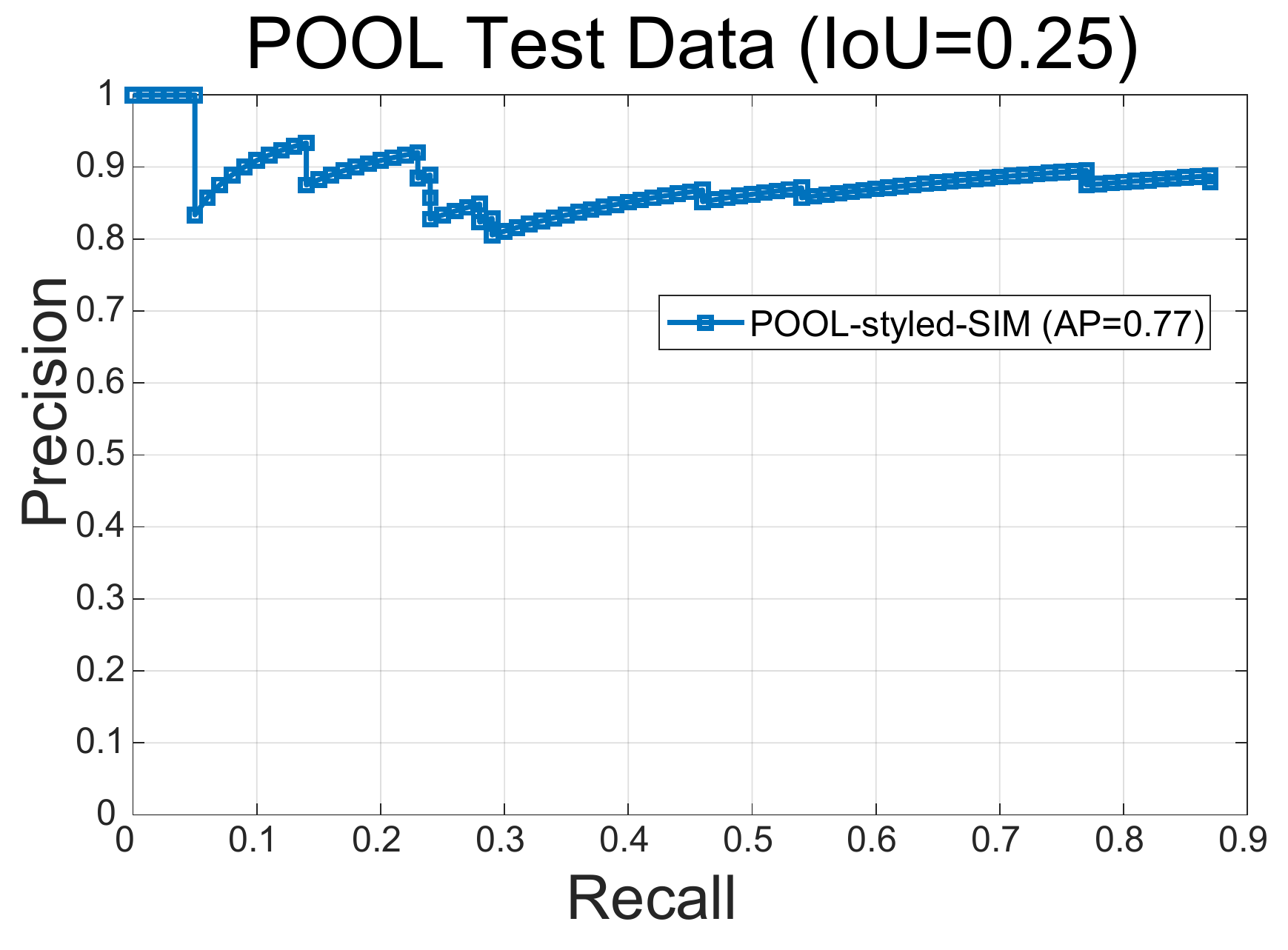}
    \label{fig:pr_POOL_style}
  }
  \subfigure[\texttt{SIM-SEA2017} to \texttt{SEA2017}]{
    \includegraphics[width = 0.45\columnwidth]{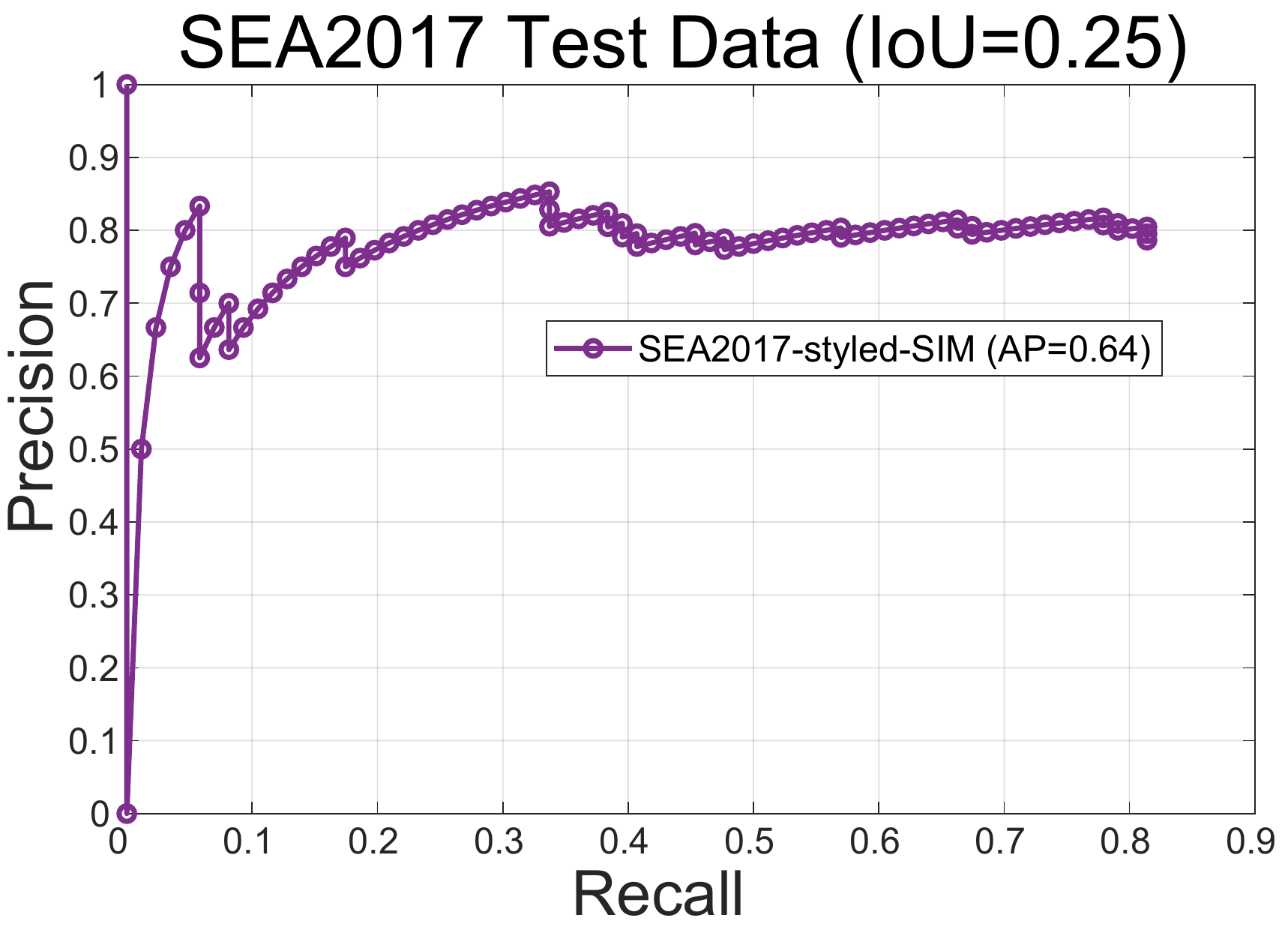}
    \label{fig:pr_SEA_style}
  }

  \caption{Object detection performance when target environment changes.
  \subref{fig:pr_POOL_style} the network is trained from simulator-generated
  images applied with water tank style, and is tested with real sonar images
  collected from water tank, \subref{fig:pr_SEA_style} the network is trained
  from simulator-generated images applied with sea style and is tested with real sea sonar images.}

	\label{fig:pr_style}
\end{figure}

Style transfer and object detection training was performed running on one NVIDIA
GTX 1080. Adam optimizer was used. The learning rates were set to $10^{-3}$ with
an exponential decay. Weight decay, $\beta_{1}$ and $\beta_{2}$ were set to
$10^{-5}$, $0.9$ and $0.999$, respectively.



We considered detection \ac{IOU} larger than 0.25 as the correct detection. For
terrestrial images, IOU $=0.5$ is often used.  Considering sonar images
resolution and underwater navigation accuracy, we alleviated criteria of the
detection IOU. We think, however, if the targeting sonar images are high
resolution such as \ac{SAS} different \ac{IOU} can be used as the detection
criterion.

\subsection{Style Transfer Performance}

We first validated the effect of the style transfer on performance. By using a
style bank, multiple aspects of the images can be synthesized. Using the base
input image from the simulator, synthetic images were generated for
\texttt{POOL} and \texttt{SEA2017}. The style transferred images are as given in
\figref{fig:style}. As can be seen in the figure, the original color map images
are style transferred to water tank and sea styles.  The style transfer results
by epoch are also given in \figref{fig:style_epoch}.  The chosen target object
evolves to be a cleaner, stronger object as the epoch increases.

We also validates the performance of the style transfer when generating and
testing for two different target environments. Using simulator-created images,
we style transferred to water tank style and sea style. Original 370
\texttt{SIM-POOL} and 370 \texttt{SIM-SEA2017} were trained with their augmented
images and tested over 735 \texttt{POOL} and 1045 \texttt{SEA2017} images.
These style transferred images from each environment were then trained and
tested with real data from each case, as in \figref{fig:pr_style}.  Both test
cases present training from styled images resulting in meaningful object
detection performance. Average precision of $0.77$ for \texttt{POOL} and $0.63$
for \texttt{SEA2017} are achieved. The accuracy when testing in a water tank is
higher than when testing at sea. This is because the noise induced from the
background sediment is lower when testing in a water tank, as can be seen in
sample images in \figref{fig:samples}.


\subsection{Simulation Training Evaluation}

\begin{figure}[!t]
	\centering
	\includegraphics[width = 0.8\columnwidth]{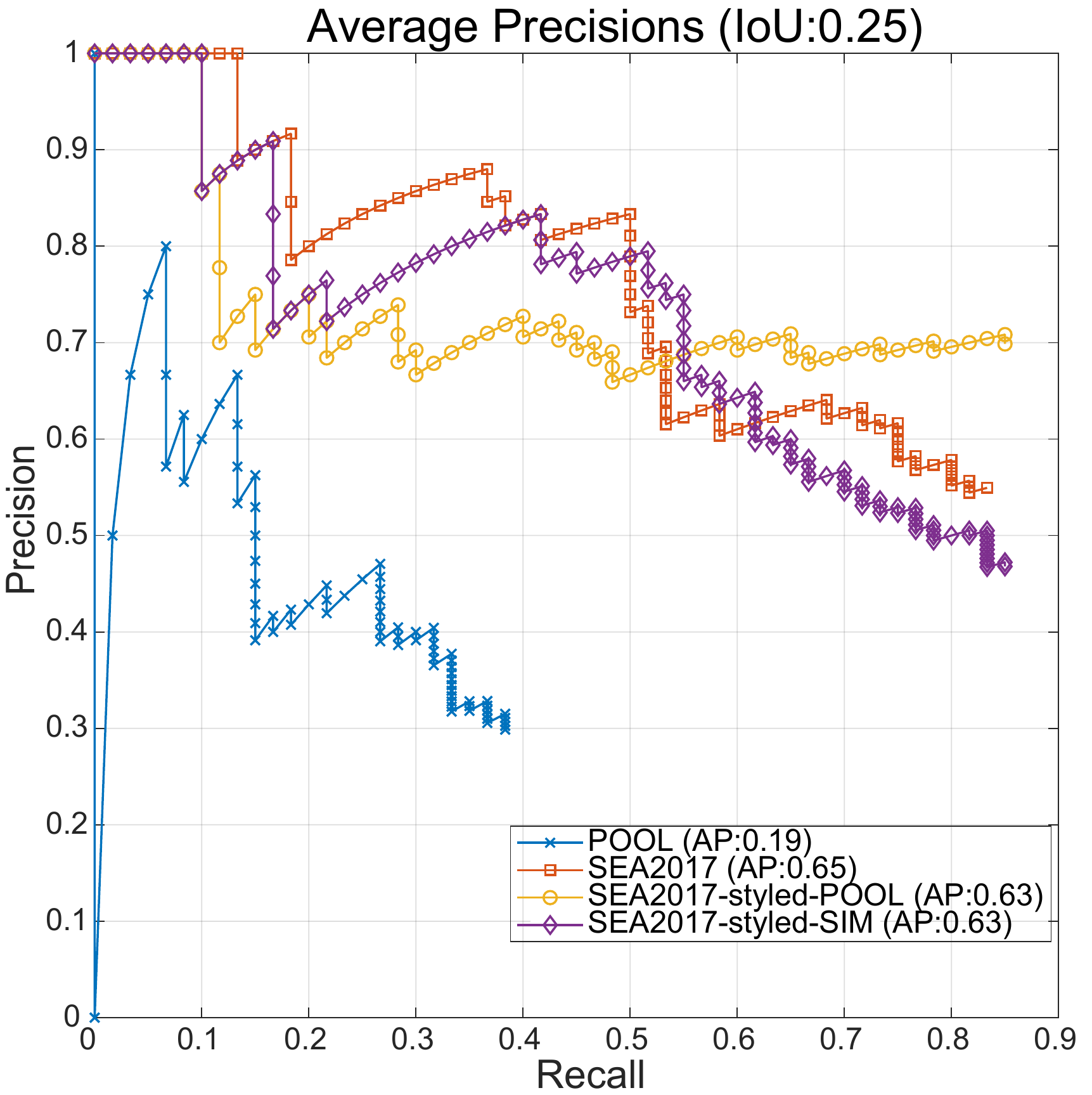}

  \caption{PR curves comparison when the network is trained by images from the
  water tank (\texttt{POOL}), style transferred images from the water tank
  (\texttt{styled-POOL}), style transferred images from the simulator
  (\texttt{styled-SIM}). Baseline result is obtained by training from real sea
  images captured in 2017 (\texttt{SEA2017}). All four cases are tested by using
  real sea sonar images captured in 2018 (\texttt{SEA2018}).}

	\label{fig:pr_curve}
\end{figure}

\begin{figure*}[!t]
  \centering
	\subfigure[Sample images]{%
    \includegraphics[width = 0.95\textwidth]{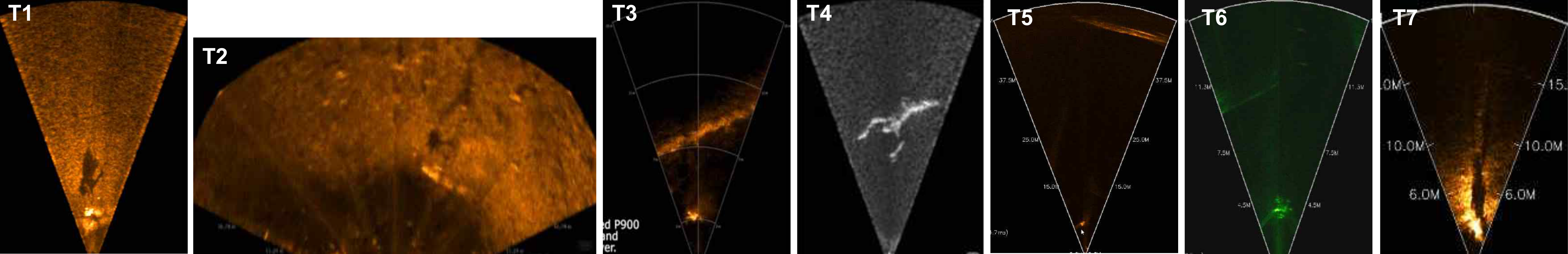}
    \label{fig:co_samples_img}
  }\\
	\begin{minipage}[]{0.9\textwidth}
		\centering
	  \subfigure[Dataset lists and video sample image description]{
	  \label{tab:co_samples}
	  \footnotesize
			\begin{tabular}{c|ccccc}
	      Name  & manufacturer					  & Image \# & Target object                 & Range [m] & Sonar type\\
				\hline\hline
				T1	  & Teledyne (P900-45) 		  & 5        & Diver standing sea floor      & 5	       & Multibeam imaging sonar\\
	      T2	  & Teledyne (P900-130) 	  & 5        & Diver swimming near sea floor & 10        & Multibeam imaging sonar\\
	      T3		& Teledyne (P900-45) 		  & 5        & Diver swimming far    			   & 10        & Multibeam imaging sonar\\
				T4		& Teledyne (P900-45) 		  & 5 			 & Diver swimming near           & 2         & Multibeam imaging sonar\\
				T5		& SonarTech         	  	& 10			 & Diver approaching to sensor   & 1-25      & Multibeam imaging sonar\\
				T6		& SonarTech             	& 10			 & Diver swimming in-Water       & 10        & Multibeam imaging sonar\\
				T7    & SonarTech       			  & 5				 & Diver standing sea floor      & 3         & Multibeam imaging sonar \vspace{5px}\\
			\end{tabular}
		\normalsize
		}
  \end{minipage}

  \caption{Test sonar images captured from company provided sample videos. T1-T4
  were sampled from videos available from Teledyne and T5-T7 were captured from
  video provided by SonarTech. \subref{fig:co_samples_img} Sample images
  from each dataset. \subref{tab:co_samples} Summary of the dataset.}

  \label{fig:co_samples}
\end{figure*}

\begin{figure}[!h]
  \centering
  \includegraphics[width = 0.95\columnwidth]{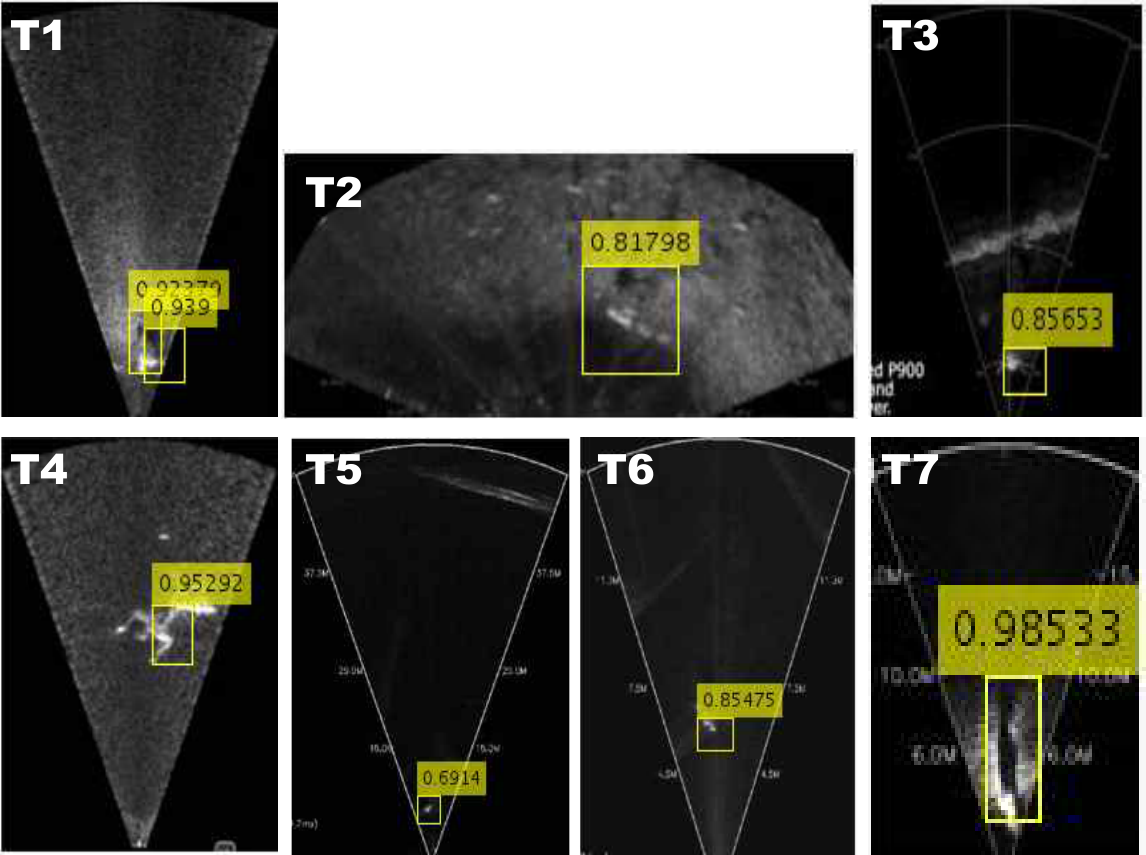}

  \caption{Test results from sample images captured from video.}

  \label{fig:video_results}
\end{figure}


If possible, training from the real sea and testing with real sea images would
be ideal. Hence, we use the object detection results trained from
\texttt{SEA2017} and tested them on \texttt{SEA2018} as the baseline,
considering that this would be the optimal training method. As can be seen in
\figref{fig:pr_curve}, the baseline provides around 0.65 accuracy when detecting
the object.

In comparison to this baseline, we performed object detection from three cases:
when trained from a water tank (735 images from \texttt{POOL}), when trained
from stylized images from a water tank  (370 images from \texttt{SIM-POOL}) and
when trained from simulator using style transfer (370 images from
\texttt{SIM-SEA2017}). The precision-recall curve comparison is provided in
\figref{fig:pr_curve}. The accuracy and detection performance are slightly
degraded compared to training from real sea data. On the other hand, the
proposed method elaborated the simulation-generated images to include
characteristics of the real sea images via style transfer. The resulting object
detection performance is comparable to that of the detection result when trained
with real sea images.

\subsection{Validation to Public Data}
\label{sec:company}

Lastly, we verified that the proposed method is applicable to other types of
sonar from two different manufacturers by testing in the various environments.
Again, we trained the network using the simulator-generated images and by
applying style transfer. As described in \figref{fig:co_samples}, we collected
sample images from various sample videos. These images contain either a standing
or swimming diver at various ranges. The sample data were collected using
different sensors and from different sediment conditions. An object's relative
size within an image varies when captured at close (T4) vs far range (T5).
Depending on the viewing angle and diver's posture, a strong shadow occurred
when the diver was standing on the sea floor (T1 and T7). When the target is
swimming in water, the ground appears separately, as in T3 and T5.

Sample test results are shown in \figref{fig:video_results}. Despite the variety
of sample cases, the target object (i.e., diver in the sea) was successfully
detected. One notable case was found in T5 when a diver approached the sonar
starting from \unit{25}{m} away from the sensor. As can be seen in the sample
and result cases, only a couple of pixels indicate the object. The learned
network suffered from this subtle information and detected the object only when
the range became closer (less than \unit{5}{m}). Also, when the target object
was found in multiple pixels within a short range, the object was found multiple
times when diver motion was greater. The motion could be highly diverse when the
diver was swimming and this level of ambiguity was well secured by the training.
Furthermore, the trained network was not fooled by other objects such as rocks
or the ground, which also appear as bright objects in the scene.

\section{Conclusion}
\label{sec:conclusion}

In this paper, we applied \ac{CNN}-based underwater object detection from sonar
images. The main objective was to overcome data limitations in the underwater
environment by synthesizing sonar images obtained from a simulator and testing
over sonar images captured in a real underwater environment. Our results
validate that the proposed image synthesizing mimics real underwater images
without actual performing dives. The proposed training solution is applicable
for various target detection by using a 3D model of the target from the
simulator.






\bibliographystyle{IEEEtranN}
\bibliography{string-long,references}

\begin{thebibliography}{30}
\providecommand{\natexlab}[1]{#1}
\providecommand{\url}[1]{#1}
\csname url@samestyle\endcsname
\providecommand{\newblock}{\relax}
\providecommand{\bibinfo}[2]{#2}
\providecommand{\BIBentrySTDinterwordspacing}{\spaceskip=0pt\relax}
\providecommand{\BIBentryALTinterwordstretchfactor}{4}
\providecommand{\BIBentryALTinterwordspacing}{\spaceskip=\fontdimen2\font plus
\BIBentryALTinterwordstretchfactor\fontdimen3\font minus
  \fontdimen4\font\relax}
\providecommand{\BIBforeignlanguage}[2]{{%
\expandafter\ifx\csname l@#1\endcsname\relax
\typeout{** WARNING: IEEEtranN.bst: No hyphenation pattern has been}%
\typeout{** loaded for the language `#1'. Using the pattern for}%
\typeout{** the default language instead.}%
\else
\language=\csname l@#1\endcsname
\fi
#2}}
\providecommand{\BIBdecl}{\relax}
\BIBdecl

\bibitem[Cho et~al.(2015)Cho, Gu, Joe, Asada, and Yu]{cho-2015}
H.~Cho, J.~Gu, H.~Joe, A.~Asada, and S.-C. Yu, ``Acoustic beam profile-based
  rapid underwater object detection for an imaging sonar.'' \emph{Journal of
  Marine Science and Technology}, vol.~20, no.~1, pp. 180--197, Mar 2015.

\bibitem[Purcell et~al.(2011)Purcell, Gallo, Packard, Dennett, Rothenbeck,
  Sherrell, and Pascaud]{purcell-2011}
M.~Purcell, D.~Gallo, G.~Packard, M.~Dennett, M.~Rothenbeck, A.~Sherrell, and
  S.~Pascaud, ``Use of remus 6000 auvs in the search for the air france flight
  447,'' in \emph{Proceedings of the {IEEE}/{MTS} {OCEANS} Conference and
  Exhibition}, Sept 2011, pp. 1--7.

\bibitem[Reed et~al.(2003)Reed, Petillot, and Bell]{reed-2003}
S.~Reed, Y.~Petillot, and J.~Bell, ``An automatic approach to the detection and
  extraction of mine features in sidescan sonar,'' \emph{IEEE Journal of
  Oceanic Engineering}, vol.~28, no.~1, pp. 90--105, Jan 2003.

\bibitem[Belcher and Lynn(2000)]{belcher-2000}
E.~O. Belcher and D.~C. Lynn, ``Acoustic near-video-quality images for work in
  turbid water,'' \emph{Proceedings of Underwater Intervention}, vol. 2000,
  2000.

\bibitem[Lee et~al.(2013)Lee, Kim, and Choi]{lee-2013}
Y.~Lee, T.~G. Kim, and H.~T. Choi, ``Preliminary study on a framework for
  imaging sonar based underwater object recognition,'' in \emph{2013 10th
  International Conference on Ubiquitous Robots and Ambient Intelligence
  (URAI)}, Oct 2013, pp. 517--520.

\bibitem[Williams and Groen(2011)]{williams-2011}
D.~P. Williams and J.~Groen, ``A fast physics-based, environmentally adaptive
  underwater object detection algorithm,'' in \emph{Proceedings of the
  {IEEE}/{MTS} {OCEANS} Conference and Exhibition}, June 2011, pp. 1--7.

\bibitem[Galceran et~al.(2012{\natexlab{a}})Galceran, Djapic, Carreras, and
  Williams]{galceran-2012}
E.~Galceran, V.~Djapic, M.~Carreras, and D.~P. Williams, ``A real-time
  underwater object detection algorithm for multi-beam forward looking sonar,''
  \emph{IFAC Proceedings Volumes}, vol.~45, no.~5, pp. 306--311, 2012.

\bibitem[Lee(2017)]{lee-2017}
S.~Lee, ``Deep learning of submerged body images from 2d sonar sensor based on
  convolutional neural network,'' in \emph{Underwater Technology (UT), 2017
  IEEE}, 2017, pp. 1--3.

\bibitem[Shin et~al.(2015)Shin, Lee, Choi, and Kim]{yshin-2015-oceans}
Y.-S. Shin, Y.~Lee, H.-T. Choi, and A.~Kim, ``Bundle adjustment from sonar
  images and {SLAM} application for seafloor mapping,'' in \emph{Proceedings of
  the {IEEE}/{MTS} {OCEANS} Conference and Exhibition}, Washington, DC, Oct.
  2015, pp. 1--6.

\bibitem[Johnnsson et~al.(2010)Johnnsson, Kaess, Englot, Hover, and
  Leonard]{johnnsson-2010}
H.~Johnnsson, M.~Kaess, B.~Englot, F.~Hover, and J.~J. Leonard, ``Imaging
  sonar-aided navigation for autonomous underwater harbor surveillance,'' in
  \emph{Proceedings of the {IEEE}/{RSJ} International Conference on Intelligent
  Robots and Systems}, 2010.

\bibitem[Inc.(2018)]{navigator}
\BIBentryALTinterwordspacing
S.~M.~T. Inc., ``Navigator,'' 2018. [Online]. Available:
  \url{http://www.sharkmarine.com/}
\BIBentrySTDinterwordspacing

\bibitem[Galceran et~al.(2012{\natexlab{b}})Galceran, Djapic, Carreras, and
  Williams]{enric-2012}
E.~Galceran, V.~Djapic, M.~Carreras, and D.~P. Williams, ``A real-time
  underwater object detection algorithm for multi-beam forward looking sonar,''
  \emph{IFAC Proceedings Volumes}, vol.~45, no.~5, pp. 306 -- 311, 2012.

\bibitem[Zhou and Chen(2004)]{zhou-2004}
X.~Zhou and Y.~Chen, ``Seafloor sediment classification based on multibeam
  sonar data,'' \emph{Geo-spatial Information Science}, vol.~7, no.~4, pp.
  290--296, 2004.

\bibitem[Williams(2015)]{williams-2015-tgrs}
D.~P. Williams, ``Fast unsupervised seafloor characterization in sonar imagery
  using lacunarity,'' \emph{{IEEE} Transactions on Geoscience and Remote
  Sensing}, vol.~53, no.~11, pp. 6022--6034, 2015.

\bibitem[Zhu et~al.(2017)Zhu, Isaacs, Fu, and Ferrari]{pzhu-2017-cdc}
P.~Zhu, J.~Isaacs, B.~Fu, and S.~Ferrari, ``Deep learning feature extraction
  for target recognition and classification in underwater sonar images,'' in
  \emph{Proceedings of the {IEEE} Conference on Decision and Control}, 2017,
  pp. 2724--2731.

\bibitem[Williams(2016)]{williams-2016-icpr}
D.~P. Williams, ``Underwater target classification in synthetic aperture sonar
  imagery using deep convolutional neural networks,'' in \emph{Proceedings of
  the International Conference Pattern Recognition}, Dec 2016, pp. 2497--2502.

\bibitem[Kim et~al.(2016)Kim, Cho, Pyo, Kim, and Yu]{jkim-syu-2016-oceans}
J.~Kim, H.~Cho, J.~Pyo, B.~Kim, and S.-C. Yu, ``The convolution neural network
  based agent vehicle detection using forward-looking sonar image,'' in
  \emph{Proceedings of the {IEEE}/{MTS} {OCEANS} Conference and Exhibition},
  2016, pp. 1--5.

\bibitem[McKay et~al.(2017)McKay, Gerg, Monga, and Raj]{jmckay-2017-oceans}
J.~McKay, I.~Gerg, V.~Monga, and R.~G. Raj, ``What's mine is yours: Pretrained
  {CNN}s for limited training sonar {ATR},'' in \emph{Proceedings of the
  {IEEE}/{MTS} {OCEANS} Conference and Exhibition}, 2017, pp. 1--7.

\bibitem[Denos et~al.(2017)Denos, Ravaut, Fagette, and Lim]{kdenos-2017-oceans}
K.~Denos, M.~Ravaut, A.~Fagette, and H.~Lim, ``Deep learning applied to
  underwater mine warfare,'' in \emph{Proceedings of the {IEEE}/{MTS} {OCEANS}
  Conference and Exhibition}, 2017.

\bibitem[Chen and Summers(2016)]{jchen-2017-jasa}
J.~L. Chen and J.~E. Summers, ``Deep neural networks for learning
  classification features and generative models from synthetic aperture sonar
  big data,'' \emph{The Journal of the Acoustical Society of America}, vol.
  140, 2016.

\bibitem[Dhurandher et~al.(2008)Dhurandher, Misra, Obaidat, and
  Khairwal]{uwsim}
S.~K. Dhurandher, S.~Misra, M.~S. Obaidat, and S.~Khairwal, ``Uwsim: A
  simulator for underwater sensor networks.'' \emph{Simulation}, vol.~84,
  no.~7, pp. 327--338, 2008.

\bibitem[Gwon et~al.(2017)Gwon, Kim, Kim, Park, Kim, and Kim]{dgwon-2017-urai}
D.-H. Gwon, J.~Kim, M.~H. Kim, H.~G. Park, T.~Y. Kim, and A.~Kim, ``Development
  of a side scan sonar module for the underwater simulator,'' in
  \emph{Proceedings of the International Conference on Ubiquitous Robots and
  Ambient Intelligence}, Jeju, S. Korea, Aug. 2017, pp. 662--665.

\bibitem[Chen et~al.(2017)Chen, Yuan, Liao, Yu, and Hua]{stylebank}
D.~Chen, L.~Yuan, J.~Liao, N.~Yu, and G.~Hua, ``Stylebank: An explicit
  representation for neural image style transfer,'' \emph{CoRR}, vol.
  abs/1703.09210, 2017.

\bibitem[Johnson et~al.(2016)Johnson, Alahi, and Fei-Fei]{jjohnson-2016-eccv}
J.~Johnson, A.~Alahi, and L.~Fei-Fei, ``Perceptual losses for real-time style
  transfer and super-resolution,'' in \emph{Proceedings of the European
  Conference on Computer Vision}.\hskip 1em plus 0.5em minus 0.4em\relax
  Springer, 2016, pp. 694--711.

\bibitem[Fan et~al.(2017)Fan, Lyu, Ying, and Hu]{yfan-2017-nips}
Y.~Fan, S.~Lyu, Y.~Ying, and B.-G. Hu, ``Learning with average top-k loss,'' in
  \emph{Advances in Neural Information Processing Systems Conference}, Long
  beach, USA, Nov. 2017.

\bibitem[Goodfellow et~al.(2014)Goodfellow, Pouget-Abadie, Mirza, Xu,
  Warde-Farley, Ozair, Courville, and Bengio]{igoodfellow-2014-nips}
I.~J. Goodfellow, J.~Pouget-Abadie, M.~Mirza, B.~Xu, D.~Warde-Farley, S.~Ozair,
  A.~Courville, and Y.~Bengio, ``Generative adversarial networks,'' in
  \emph{Advances in Neural Information Processing Systems Conference},
  Montreal, CANADA, Nov. 2014.

\bibitem[Ren et~al.(2017)Ren, He, Girshick, and Sun]{ren2017faster}
S.~Ren, K.~He, R.~Girshick, and J.~Sun, ``Faster r-cnn: towards real-time
  object detection with region proposal networks,'' \emph{IEEE Transactions on
  Pattern Analysis \& Machine Intelligence}, no.~6, pp. 1137--1149, 2017.

\bibitem[Zitnick and Doll{\'a}r(2014)]{zitnick2014edge}
C.~L. Zitnick and P.~Doll{\'a}r, ``Edge boxes: Locating object proposals from
  edges,'' in \emph{European conference on computer vision}, 2014, pp.
  391--405.

\bibitem[Uijlings et~al.(2013)Uijlings, Van De~Sande, Gevers, and
  Smeulders]{uijlings2013selective}
J.~R. Uijlings, K.~E. Van De~Sande, T.~Gevers, and A.~W. Smeulders, ``Selective
  search for object recognition,'' \emph{International journal of computer
  vision}, vol. 104, no.~2, pp. 154--171, 2013.

\bibitem[Girshick(2015)]{girshick2015fast}
R.~Girshick, ``Fast r-cnn,'' in \emph{Proceedings of the IEEE international
  conference on computer vision}, 2015, pp. 1440--1448.

\end{thebibliography}


\end{document}